# Event Structure of Transitive Verb: A MARVS perspective

Jia-Fei Hong[1] Kathleen Ahrens[2] Chu-Ren Huang[3]

[1] Department of Applied Chinese Language and Culture, National Taiwan Normal University, Taiwan

[2] Language Centre, Hong Kong Baptist University, Hong Kong

[3] Faculty of Humanities, The Hong Kong Polytechnic University, Hong Kong

jiafeihong@ntnu.edu.tw; ahrens@hkbu.edu.hk; churen.huang@polyu.edu.hk

**Abstract**: Module-Attribute Representation of Verbal Semantics (MARVS) is a theory of the representation of verbal semantics that is based on Mandarin Chinese data (Huang et al. 2000). In the MARVS theory, there are two different types of modules: Event Structure Modules and Role Modules. There are also two sets of attributes: Event-Internal Attributes and Role-Internal Attributes, which are linked to the Event Structure Module and the Role Module, respectively. In this study, we focus on four transitive verbs as *chi1* "eat", *wan2* "play", *huan4* "change" and *shao1* "burn" and explore their event structures by the MARVS theory.

## 1. Introduction

In this sense prediction study, four verbs—*chi1* "eat", *wan2* "play", *huan4* "change", and *shao1* "burn"—have been selected as target words in order to predict their senses. There are two main reasons these four target words have been chosen: (1) they are all transitive verbs; and (2) they each have more than two senses. These two reasons may not fully explain why they have been selected from the many transitive verbs available in the corpus. Therefore, to clarify further our reasoning in selecting these four target words, we will employ the Module-Attribute Representation of Verbal Semantics Theory (Huang et al. 2000).

Before detailing why these four target verbs were selected, we will introduce the Module-Attribute Representation of Verbal Semantics Theory (MARVS Theory).

Next, we will discuss these four target words in relation to the MARVS Theory. Finally, we would like to discuss the four transitive verbs as *chi1* “eat”, *wan2* “play”, *huan4* “change” and *shao1* “burn” and explore their event structures by the MARVS theory.

## 2. Module-Attribute Representation of Verbal Semantics

Module-Attribute Representation of Verbal Semantics (MARVS) is a theory of the representation of verbal semantics that is based on Mandarin Chinese data (Huang et al. 2000). In the MARVS theory, there are two different types of modules: Event Structure Modules and Role Modules. There are also two sets of attributes: Event-Internal Attributes and Role-Internal Attributes, which are linked to the Event Structure Module and the Role Module, respectively.

In the MARVS theory, Huang et al. (2000) mentioned that lexical knowledge is classified into two types: (1) structural information, which is represented by means of the composition of atomic modules; and (2) content information, which is represented by means of attributes attached to these modules. In addition, the roles that participate in the event are represented in the Role Modules. The semantic attributes pertaining to the complete event are called the Event-Internal Attributes, which are attached to the Event Structure Module. In addition, Event-Internal Attributes refer to the semantics of the event itself. Moreover, the semantic attributes pertaining to each role are termed Role-Internal Attributes, which are attached to the appropriate role within the Role Module. The overall shape of the Event Structure Module is defined by the composition of the five Event Modules. It is important to note that the eventive information is attached to the sense of a verb. Verbs with different senses will have different eventive information. The representation of the MARVS theory is shown below in Figure 1:

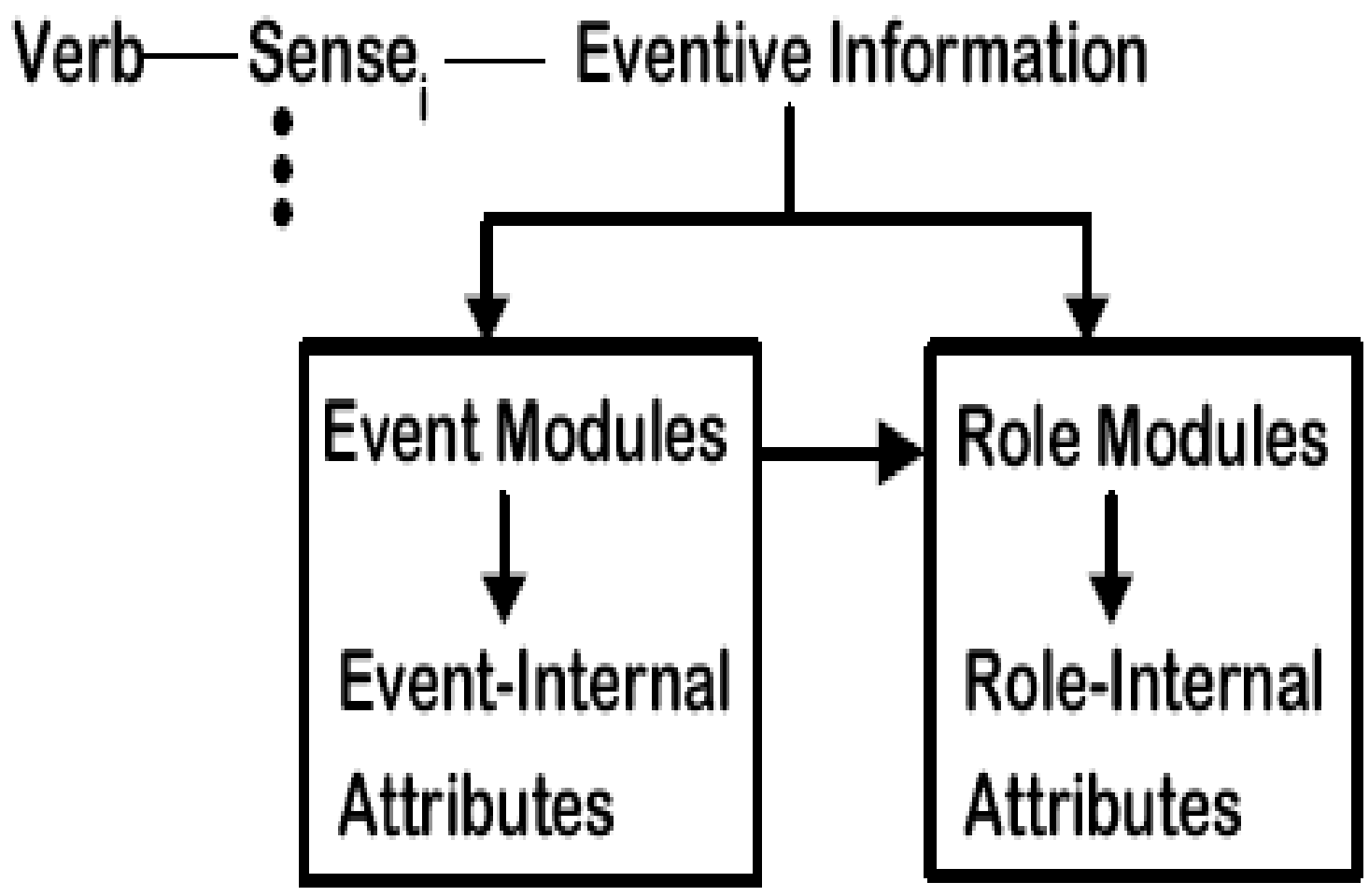

Figure 1: Module-Attribute Representation of Verbal Semantics (MARVS)

A central issue in lexical semantics is the representation of events (e.g., Jackendoff 1983 and Pustejovsky 1991). In particular, Huang et al. (2000) proposed the MARVS theory in which event structures can be created from a small set of event modules, while the backbone of verbal semantics can be considered as combinations of these event modules. Event modules are the building blocks of linguistic event structures, and they can be used in combination or alone. When used alone, they are atomic logical event structures. Huang et al. (2000) showed these five atomic event structures and gave them short definitions as shown below in (1) through (5):

*Atomic logical event structures*

(1) ● Boundary (includes a Complete Event)

Boundary: an event module that can be identified by means of a temporal point and must be regarded as a whole.

(2) / Punctuality

Punctuality: an event module that represents a single occurrence of an activity that cannot be measured based on duration.

(3) ///// Process

Process: an event module that represents an activity that has a time course that can be measured in terms of its temporal duration.

(4) ▁ State

State: a homogeneous event module in which the concept of temporal duration is irrelevant. It is neither punctual nor does it have a time course.

(5) ^^^ Stage

Stage: an event module consisting of iterative sub-events.

In other words, Huang et al. (2000) postulated that these five atomic event structures are the only building blocks necessary to capture the range of complex linguistic event structures.

Regarding the Role Modules in the MARVS theory, Huang et al. (2000) mentioned that they contain the focused roles of an event and typically include all required arguments but can also include optional arguments and adjuncts. Therefore, regarding the roles, they considered the following as some of these arguments: Agent, Cause, Causer, Comparison, Experiencer, Goal, Instrument, Incremental Theme, Location, Locus, Manner, Range, Recipient, Source, Target, Theme, etc. They also explained how their Role Module works with an optional argument. Moreover, Huang et al. (2000) pointed out the Role-Internal Attributes that interact with context-induced meaning to determine the appropriate reading while selectional restrictions are projected from a fixed lexical entry.

## 3. Previous Studies

Various studies of verbal semantics from the MARVS perspective have been conducted, including Ahrens et al. (2003), Hong et al. (2008), and Chung and Ahrens (2008). Ahrens et al. (2003) examined this theory in light of English data and Mandarin Chinese data. Hong et al. (2008) took the MARVS theory to explicate different levels of representations for event selection and coercion of two verbs of ingestion. Hong et al. (2008) were also able to establish a model of event-type selection and coercion, where they could predict the meaning of a non-typical event-type object, as well as predict metaphoric meanings and Event-Module Attributes. Chung and Ahrens (2008) suggested that the operational steps underlying a MARVS analysis could be improved by analyzing the sense distribution of the near-synonyms and by looking at the Mutual Information (MI) values of the collocating words.

In the Ahrens et al. (2003) study, they looked at the near-synonym contrast of the verbs “put” and “set” based on data from the sampler of the British National Corpus. Moreover, they also examined the distributional differences that exist for *bai3* and *fang4* in Mandarin Chinese.

The English verbs “put” and “set” seem synonymous and interchangeable in most contexts as shown in examples (6) through (8b) below:

(6) MARVS representation of *bai3* and *fang4*

*bai3* ●▁ <Agent, Theme, **Location**>
|
[design]

*fang4* ●▁ < Agent, Theme, Location >

(7a) Put/set the book on the table.

(7b) He set/put the pin on the cushion.

(8a) ta zhengzai fang/*bai shu zai zhuo-shang

s/he DURATIVE put/set book at table-top

(8b) She is putting/setting the book on the table.

From this data analysis, Ahrens et al. (2003) found that the near-synonym pair "put" and "set", or *bai3* and *fang4,* have nearly complimentary distributions, which clearly indicates their semantic contrast from the MARVS perspective.

In the case of Hong et al. (2008), they mentioned that event semantics in general and event-type coercion in particular, have been a challenging yet rewarding topic in verbal semantics (Pustejovsky 1995). However, there have been few corpus-based empirical accounts discussing the range of event-type coercions based on the lexical meanings of the verbs. In their paper, Hong et al. (2008) explored the possible types of event coercions for two verbs of ingestion in Mandarin Chinese. In particular, they showed that the different types of coercions could be predicted by the bifurcation of Event-Internal Attributes and Role-Internal Attributes proposed in the MARVS theory (Huang et al. 2000). Data examined were taken from the Chinese Gigaword Corpus and accessed through Chinese Word Sketch.

The MARVS theory offers a straightforward way to account for the three different types of event coercions. The two types of modules in MARVS, Event Structure Modules and Role Modules allow the description of two sets of attributes, namely Event-Internal Attributes and Role-Internal Attributes. For example, intuitively, *chi*1 “eat” and *he1* “drink” select the [+solid] feature and the [+liquid] feature, respectively, as Role-Internal Attributes.

Moreover, Chung and Ahrens (2008) proposed two additional criteria for near-synonym analyses and suggested that these criteria further allow the operationalization of the steps used to identify contrasts in near-synonyms. In addition, they proposed that analysis of sense distribution and MI values can be used to state the differences between two nearly synonymous verbs.

In this verbal event structure study, we will utilize the MARVS theory to examine whether the four target words--- *chi1* “eat”, *wan2* “play”, *huan4* “change”, and *shao1* “burn” belong to the same verbal category.

## 4. Empirical Data Collection

Explanations of Chinese Gigaword Corpus and Chinese Word Sketch (CWS) can be found in Kilgarriff et al. (2005), Huang et al. (2005), Ma and Huang (2006) and Hong and Huang (2006). CWS is a combination of the Chinese GigaWord Corpus and the linguistic search tool, Word Sketch Engine (Kilgarriff et al., 2004), a very powerful tool for extracting meaningful grammatical relations given a sufficiently large corpus (Kilgarriff et al., 2004, Huang et al., 2005).

The CWS database is collected from the Chinese Gigaword Corpus, which contains about 1.1 billion Chinese characters made up of more than 700 million characters from Taiwan’s Central News Agency and nearly 400 million characters from China’s Xinhua News Agency. The segmentation and tagging was performed automatically and included automatic and partially manual post-checking. The precision accuracy is estimated to be over 96.5% (Ma and Huang 2006).

Kilgarriff et al. (2004) developed Sketch Engine to facilitate the efficient use of gargantuan corpora. Sketch Engine (SKE, also known as the Word Sketch Engine is a novel Corpus Query System incorporating word sketches, grammatical relations, and a distributional thesaurus.

The advantage of using Sketch Engine as a query tool is that it pays attention to the grammatical context of a word rather than just looking at an arbitrary number of adjacent words. In order to test the cross-lingual robustness of Sketch Engine as well as to propose a powerful tool for collocation extraction using a large scale corpus that requires minimal pre-processing, we constructed a Chinese Word Sketch Engine (CWS) by loading the Chinese Gigaword Corpus onto the Sketch Engine platform Kilgarriff et al. (2005). All components of Sketch Engine were implemented, including *Concordance, Word Sketch, Thesaurus and Sketch Difference*.

In this study, we focus on the four target words as transitive verbs, therefore, we use query functions of Chinese Word Sketch to find their collocation nouns. Moreover, we analysis these collocation nouns of four transitive verbs by MARVS theory to divide their completed senses in order to their particular or common transitive verbal event structures.

## 5. Four Verbs in MARVS

### 5.1. Tests for the Four Verbs

Since the MARVS theory can present verbal event structures and show their logical primary units and entailments, each verb will be interpreted by its verifiable entailment. In other words, for the four target words—*chi1* “eat”, *wan2* “play”, *huan4*

“change”, and *shao1* “burn”—in this sense prediction study, we will use the MARVS theory to explain why we selected these particular words and to show their common points.

Concerning the event modules in the MARVS theory, Huang et al. (2000) pointed out the complete event and the process event of the verbs. They considered that only boundaries (including stand-alone complete events) could be identified with a temporal point as shown in example (9) below:

(9) *Sheme shihou V (le)*

When V ASP

They also considered that since a process event encodes a time course, a durational phrase naturally measures the length of the time course and can distinguish between process events and boundary/complete events as shown in example (10) below:

(10) V le Duration

V ASP Duration

a. *(yizhi pao)*

always run

b. *(yizhi pao)* *pao le* *san* *ge* *xiaoshi*

always run run ASP three CL hours

'(He has kept on) running for three hours.'

Since *chi1* "eat", *wan2* "play", *huan4* "change", and *shao1* "burn" are all verbal events that can be regarded as complete events, they also need a time course and a durational phrase. In addition, they all have an end point. We will employ examples (11) through (18) below to illustrate these four words:

(11) *Sheme shihou* *chi* *(le)*

When eat

'When did someone eat?'

(12) *Sheme shihou* *wan* *(le)*

When play

'When did someone play?'

(13) *Sheme shihou huan (le)*

When change

'When did someone change?'

(14) *Sheme shihou shao (le)*

When burn

'When did someone burn?'

(15) *(yizhi chi) chi le san ge xiaoshi*

always eat eat ASP three CL hours

'(He has kept on) eating for three hours.'

(16) *(yizhi wan) wan le san ge xiaoshi*

always play play ASP three CL hours

'(He has kept on) playing for three hours.'

(17) *(yizhi huan) huan le san ge xiaoshi*

always change change ASP three CL hours

'(He has kept on) changing for three hours.'

| | | | | | | |
|---|---|---|---|---|---|---|
| (18) | *(yizhi shao)* | *shao* | *le* | *san* | *ge* | *xiaoshi* |
| | always burn | burn | ASP | three | CL | hours |
| | ‘(He has kept on) burning for three hours.’ | | | | | |

Therefore, according to the MARVS theory, since complete and boundary events both have a delimiting temporal point, the durational phrase can only be interpreted as the distance between a reference point in time and that delimiting temporal time. Following the interpretations of the four target words in examples (11) through (18), we can easily verify their constructions as defined by the MARVS theory.

From these examples and explanations, it can be seen that when these events proceed, the agents need to begin making other preparations; therefore, the temporal points are both necessary and important. In addition, according to examples (11) through (18) above, a time course or the duration of the verbal event must exist. Logically, the boundary or the end point also must exist in order for these verbal events to be regarded as a complete event. That is to say, the initial points, time courses, and end points all exist for the verbal events of the four target words. According to the MARVS theory, the event modules for the four target words seem to indicate that they are undergoing the “*Bounded Process.*” Therefore, the representations of the Event-Internal Attributes and the Role-Internal Attributes of the

MARVS theory used to examine the four target words are shown in examples (19) through (22) below:

(19) *chi1* “eat” ● ///// ● <Agent, Theme, Goal>
| 
[nourishment]

(20) *wan2* “play” ● ///// ● <Agent, Theme, Goal>
|
[mood]

(21) *huan4* “change” ● ///// ● <Agent, Theme, Goal>
|
[replacement]

(22) *shao1* “burn” ● ///// ● <Agent, Theme, Goal>
|
[transformation]

From the above examples (19) through (22), the four target words have common Role-Internal Attributes—agent, theme, and goal—but they have different Event-Internal Attributes. For *chi1* “eat”, the goal is nourishment; for *wan2* “play”, the goal is emotion; for *huan4* “change”, the goal is replacement; and for *shao1*

“burn”, the goal is transformation. Regarding the four target words, the goals of these verbs all indicate different results and changes, which can be referred to as states, such as physical or mental and concrete or abstract.

## 5.2. Analysis for the Four Verbs

*Chi1* “eat”, for example, is a typically physical action in which the agent needs to prepare the food, put the food into his/her mouth, chew and swallow the food for nourishment, and then finish the entire event via ingestion. Looking at the complete event of *chi1* “eat”, when the agent prepares the food for ingestion, this can be regarded as the initial boundary of the complete event. When the agent puts the food into his/her mouth, and then chews and swallows the food, this can be regarded as a time course and a durational phrase, respectively. When the agent finishes putting the food into his/her mouth, chewing, and swallowing the food, this can be regarded as the end boundary of the complete event—ingestion—for *chi1* “eat”.

To illustrate this process, we will use the following example: 吃蘋果 (*chi1 ping2 guo3* “eat an apple”). If the agent wants to eat an apple, s/he needs to prepare the apple, put the apple into his/her mouth, chew and swallow the apple, and when the agent finishes eating this apple, it can be said that the apple provided nourishment to the agent. This process is shown below in example (23):

(23) The complete event of *chi1* “eat”

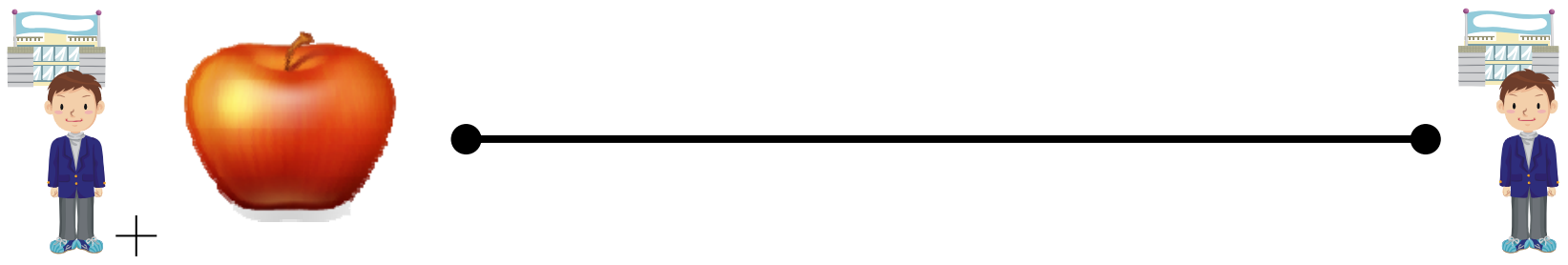

Boundary (initial point) Boundary (end point)

Apple Nourishment

In other words, when the agent ingests the food, it is reasonable to assume that the event goal is the agent obtaining nourishment. It also indicates a change in the physical state of the agent at the same time. Even when *chi1* “eat” is interpreted in a metaphorical sense, if apple is changed to any abstract object, the agent will still obtain the nourishment s/he needs. The interpretation and representation of 吃奶水 (*chi1 nai3 shui3* “eat mike”), for example, can be observed in Hong et al (2008), where they interpreted that when metaphoric uses are involved, “奶水 (*nai3 shui3* “mike”)” refers to nourishment for either the body or the soul. In addition, they also considered that 吃奶水 (*chi1 nai3 shui3* “eat mike”) can be easily represented in the MARVS theory by Subject -Internal Attributes.

Looking at the second target word—*wan2* “play”—as an example using the MARVS theory, it can be seen that it is also a complete event. According to the sense division of *wan2* “play” in CWN analysis, the original physical sense of play is an action that usually results in fun and enjoyment. In our opinion, fun and enjoyment would be classified under human mood. That is to say, when the agent plays a game, such as soccer, scoring a goal leads to the agent feeling enjoyment and pleasure. This can change the agent's mood, making the agent feel better over the course of time. When the agent finishes playing the game, s/he obtains enjoyment, the event goal. Looking at 玩足球 (*wan2 zu2 qiu2* "play soccer") as an example for *wan2* "play" in the MARVS theory, it can be regarded as a complete event because it has three event features: the beginning of the game (initial boundary), the process of playing the game (durational phrase), and the end of the game (end boundary). Example (24) below illustrates the complete event for *wan2* "play":

(24) The complete event of *wan2* "play"

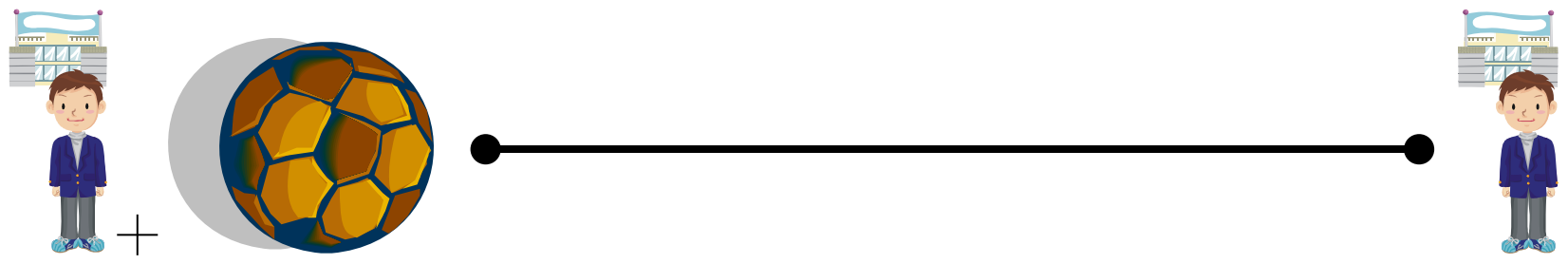

Boundary (initial point) Boundary (end point)

Soccer Mood (enjoyment)

If I employ the physical sense of the complete event of *wan2* "play" to explain its metaphorical sense using the MARVS theory, a similar condition will be interpreted. Using the example 玩網路 (*wan2 wang3 lu4* "surf the Internet"), because the agent has many interests and explores the Internet completely, the eventual goal is to obtain search results that make the agent feel pleasure and enjoyment. In other words, when the agent finishes the complete event of *wan2* “play”, the agent ultimately obtains a lovely mood.

In the case of the third target word—*huan4* “change”—the physical sense of the event is to exchange X for Y or to replace something. Therefore, the event of *huan4* “change” is also regarded as a complete event because there is an initial boundary, a duration in which the change takes place, and an end boundary via the enplanements of the event modules in the MARVS theory. Therefore, *huan4* “change” can be interpreted as 一直換 (*yi1 zhi2 huan4* “always change”) or 換了三個小時 (*huan4 le5 san1 ge4 xiao3shi2* “changing for three hours”). Example (25) below illustrates the process of changing clothes, as new clothes replace original clothes for the agent:

(25) The complete event of *huan4* “change”

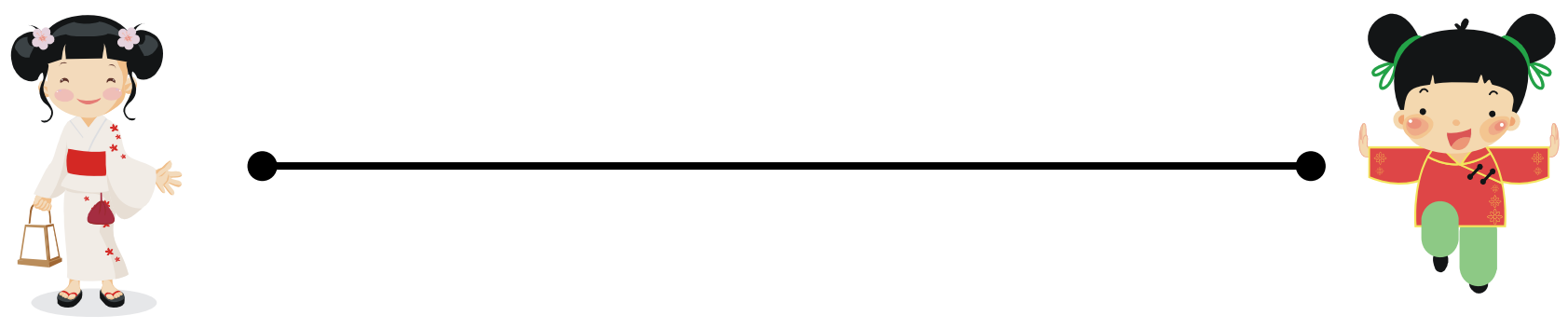

Boundary (initial point) Boundary (end point)

Original Clothes Replacement (new clothes)

Moving on to the metaphorical sense of the complete event using the MARVS theory, 搭飛機換火車 (*da1 fei1 ji1 huan4 huo3 che1* "take the airplane and transfer the train") and 賣房子換美金 (*mai4 fang2 zi5 huan4 mei3 jin1* "sell the house to change U.S. dollars"), for example, both indicate that original conditions have changed, replacing the original complete event of *huan4* "change" with new conditions.

Regarding the fourth target word—*shao1* "burn"—the conditions of the complete event are all similar to *chi1* "eat", *wan2* "play", and *huan4* "change". There are essential elements represented at the initial boundary point, during the process itself, and at the end boundary point. The initial boundary point indicates that the agent is preparing to burn something, the process indicates that the event is proceeding, and the end boundary point indicates that the event is finished and the result implies a transformation feature. Example (26) below illustrates the semantic features and elements of the complete event of *shao1* "burn":

(26) The complete event of *shao1* "burn"

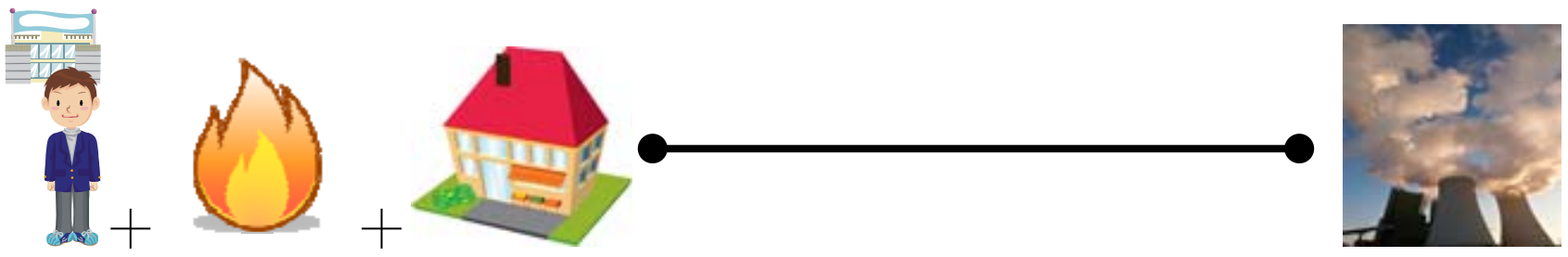

Boundary (initial point) Boundary (end point)

House Transformation (burning)

Based on the results of the other three target words, it is understood that there will be the same interpretation for the metaphorical sense of the complete event of *shao1* “burn”. Looking at the examples 燒錢 (*shao1 qian2* “use money”) and 燒油 (*shao1 you2* “use oil”), in order to achieve the goal, the agent needs to complete an action. For 燒錢 (*shao1 qian2* “use money”), the agent uses money to buy books so s/he can own the books while for 燒油 (*shao1 you2* “use oil”), the agent uses oil to make the car start so s/he can drive the car. In the complete event of *shao1* “burn”, using money to buy and own books and using oil to start the car’s engine and drive indicates that a transformation has taken place, meaning the original condition has transformed into a new condition: the agent has money and uses it to buy books, and the agent has oil and uses it to drive a car.

## 6. Summary

According to the MARVS theory, we have explained the Event-Internal Attributes of the Event Structure Modules and the Role-Internal Attributes of the Role Modules for the four target words used in this study. In addition, we presented their common points and constructions and pointed out their different internal attributes. Finally, following the verb module attribute representation of the MARVS theory, we was able to determine and explain that *chi1* “eat”, *wan2* “play”, *huan4* “change”, and

*shao1* “burn” belong to the same verbal category, and even though they display slight differences.

## Website resources